\documentclass[letterpaper,journal]{IEEEtran}
\usepackage{algorithm}
\usepackage{algpseudocode}      
\usepackage{array}
\usepackage[caption=false]{subfig}
\usepackage{textcomp}
\usepackage{stfloats}
\usepackage{url}
\usepackage{verbatim}
\usepackage{graphicx}
\usepackage[justification=centering]{caption}
\usepackage{cite}
\usepackage[utf8]{inputenc} 
\usepackage[T1]{fontenc}    
\usepackage{hyperref}       
\usepackage{url}            
\usepackage{booktabs}       
\usepackage{amsmath,amssymb,amsfonts}
\usepackage{nicefrac}       
\usepackage{doi}
\usepackage{float}
\usepackage{xcolor}
\usepackage{xparse}
\usepackage{orcidlink}
\usepackage{comment}
\usepackage{multicol,multirow}
\usepackage{bbold}

\title{Privacy-Preserving and Verifiable Approximate Distributed Coded Computing}

\author{ Xavier
   Mart\'{\i}nez-Lua\~na$^\text{\orcidlink{0000-0001-9066-983X}}$,
   Alba Gude-Santos$^\text{\orcidlink{0009-0008-7273-7593}}$, Manuel
Fern\'andez-Veiga$^\text{\orcidlink{0000-0002-5088-0881}}$~\IEEEmembership{Senior
     Member,~IEEE},
   Rebeca P. D\'{\i}az-Redondo$^\text{\orcidlink{0000-0002-2367-2219}}$~\IEEEmembership{Senior
     Member,~IEEE},  \thanks{X. Mart\'{\i}nez-Lua\~na and A. Gude-Santos are with the Galician
     Research and Development Center in Advanced Telecommunications
(GRADIANT)
     Estrada do Vilar, 56-58, Vigo, 36214, Spain. Email:
xmartinez@gradiant.org}
   \thanks{X. Mart\'{\i}nez-Lua\~na, R. P. D\'{\i}az-Redondo, and
     M. Fern\'andez-Veiga are with atlanTTic, Information \& Computing Lab (ICLAB), Telecommunication Engineering School, Universidade de Vigo Vigo, 36310, Spain. Emails: xamartinez@alumnado.uvigo.gal, rebeca@det.uvigo.es,
     mveiga@det.uvigo.es}
   \thanks{This work was partially funded by the European Union’s Horizon Europe Framework Programme for Research and Innovation Action under project TRUSTED (proj. no. 101168467), the grant DISCOVERY (PID2023-148716OB-C31) funded by MCIU/AEI/10.13039/501100011033 and FEDER, "TRUFFLES: TRUsted Framework for Federated LEarning
Systems, within the strategic cybersecurity projects (INCIBE, Spain), funded
by the Recovery, Transformation and Resilience Plan (European Union, Next
   Generation)". Additionally, it also has and it also has been funded by the Galician Regional Government under project ED431B 2024/41 (GPC).
   
   Funded by the European Union. Views and opinions expressed are however those of the authors only and do not necessarily reflect those of the European Union. Neither the European Union nor the granting authority can be held responsible for them.} }

\begin{document}

\maketitle

\begin{abstract}
Distributed machine learning enables collaborative model training without centralizing data,
but it also exposes learning processes to privacy leakage and malicious manipulation. Existing
defenses typically address these threats in isolation and are often tailored to specific learning
paradigms or model architectures, limiting their applicability in realistic deployments. In
particular, federated learning and decentralized learning exhibit distinct adversarial surfaces
that are rarely addressed within a unified framework.
In this paper, we present a model-agnostic framework for adversary-resistant distributed learning
that jointly addresses privacy preservation and malicious behavior across both federated and
decentralized settings. Our approach combines paradigm-specific defense mechanisms with GPBACC, a
privacy-enhancing coded computing technique applicable to arbitrary machine learning models. For
federated learning, we integrate robust aggregation strategies to mitigate the impact of malicious
participants, while for decentralized learning we employ approximate decode-and-compare and group
testing techniques to enable lightweight verification and adversary isolation without relying on a
trusted aggregator.
Crucially, we evaluate the proposed framework through an explicit, attack-driven analysis. We
implement representative privacy attacks and malicious behaviors, and empirically demonstrate that
the combination of GPBACC with robust aggregation and verification mechanisms significantly reduces
privacy leakage and improves resilience against active adversaries. These results suggest that
privacy-enhancing coded computing, when combined with appropriate adversary-resistance strategies,
provides a practical and deployable foundation for secure distributed machine learning.
\end{abstract}


\section{Introduction}
\label{sec:introduction}

Distributed and collaborative machine learning have become a cornerstone of modern data-driven systems, enabling multiple participants to jointly train models without centralizing raw data. Paradigms such as Federated Learning (FL) and Decentralized Learning (DL) are increasingly adopted in privacy-sensitive and large-scale environments, including healthcare, finance, and industrial systems. However, despite the inherent increase in privacy protection, neither paradigm fully avoids privacy and security issues, and both expose learning processes to privacy leakage and malicious manipulation~\cite{nguyen2023preserving, mothukuri2021survey}.

Existing protection mechanisms typically address these threats in isolation. One strategy is
to use Privacy-Enhancing Technologies (PETs), including homomorphic 
encryption~\cite{xie2024efficiency} and secure multi-party 
computation~\cite{tran2023novel}, which provide strong theoretical guarantees but often 
incur prohibitive computational and communication overheads, limiting their deployability 
in realistic learning scenarios. On the other hand, robust aggregation 
techniques~\cite{Pillutla2022} and anomaly detection methods~\cite{yang2020adversary} aim 
to mitigate poisoning and Byzantine attacks, yet they rarely address the problem of 
privacy leakage, nor are they easily transferable across different learning architectures 
and deployment models.

Moreover, most prior work focuses predominantly on federated learning~\cite{zhang2021survey} with a central aggregator, while decentralized learning~\cite{beltran2023decentralized, yuan2024decentralized}, where data is exchanged instead of model updates, remains comparatively underexplored from a unified adversarial-resilience perspective. This fundamental difference exacerbates both privacy and integrity risks, demanding fundamentally different
verification and defense mechanisms.

Consequently, practical adversary resistance in distributed learning requires a comprehensive
approach that jointly addresses privacy and malicious behavior across both learning paradigms (FL and DL), without restricting applicability to specific model architectures. To this end, we introduce a unified framework that combines complementary adversary-resistance strategies for both paradigms, enabled by our previous research work on Generalized Privacy-aware Berrut Approximated Coded Computing (GPBACC~\cite{gpbacc25}) as a PET agnostic to the AI model and the learning strategy. GPBACC is a privacy-preserving extension of Berrut Approximated Coded Computing~\cite{BACC:NA20}, a scheme from the Coded Distributed Computing domain that allows to distributedly approximate any arbitrary function while resisting stragglers. 

Built upon this technological foundation, our framework integrates robust aggregation strategies tailored to federated learning settings, as well as approximate decode-and-compare (DC) mechanisms combined with group testing techniques for decentralized learning environments. Both mechanisms are deliberately designed to address the distinct threat models inherent to each paradigm while preserving architectural generality.

Additionally, our contributions extend beyond design and integration. Rather than assuming privacy or robustness by construction, we provide a practical, attack-driven evaluation of the proposed framework. We explicitly implement and evaluate representative privacy attacks and malicious behaviors, empirically demonstrating how the combination of GPBACC with paradigm-specific adversary-resistance mechanisms achieves both privacy preservation and resilience against active adversaries. Thus, the main contributions of our work can be summarized as follows:
\begin{itemize}
    \item Definition of a unified adversary-resilient framework for distributed learning that jointly addresses privacy leakage and malicious behavior across both federated learning and decentralized learning settings.
    \item Specifically for federated learning architectures, we propose combining GPBACC with robust aggregation strategies, while preserving its model-agnostic nature.
    \item Specifically for decentralized learning architectures, we propose complementing
    GPBACC with (i) approximate decode-and-compare techniques, and (ii) group testing techniques, while preserving its model-agnostic nature.
    \item Evaluation of the global framework for both privacy preservation and resistance to malicious participants using practical attack-based tests.
\end{itemize}

To sum up, by bridging privacy enhancement and adversarial robustness across learning paradigms, our proposal contributes toward deployable, architecture-independent defenses for real-world distributed machine learning systems.

The structure of the paper is as follows. After summarizing and critically reviewing the state of the art (Sect.~\ref{sec:related_work}), we provide a brief overview of the technical background required to introduce our proposal (Sect.~\ref{sec:prel}). The threat model assumptions are detailed in Sect.~\ref{sec:threat-model}, while the mechanisms designed to enhance adversary resistance are presented in Sect.~\ref{sec:gpbacc-adversary-resistance}. Finally, we thoroughly discuss the validation results (Sect.~\ref{sec:results}) and conclude with a brief section on conclusions and future work (Sect.~\ref{sec:conclusiones}).

\section{Related Work}
\label{sec:related_work}

This work lies at the intersection of privacy-preserving machine learning, adversarially robust distributed learning, and coded computing. Accordingly, we organize the related literature along four main research directions. First, we review privacy-enhancing technologies for distributed learning, which aim to protect sensitive information during model training. Second, we examine prior work on robust aggregation mechanisms in federated learning, designed to mitigate the impact of faulty or malicious participants. Third, we discuss approaches addressing adversary-resistant decentralized learning, with a focus on resilience against adversarial behaviors in fully distributed settings. Finally, we survey recent advances in coded computing techniques applied to machine learning, which seek to improve robustness and efficiency in distributed training. 

Although we include a brief comparison between existing state-of-the-art approaches and our proposed solution, it is worth emphasizing that this paper introduces a unified, attack-validated framework that combines model-agnostic privacy-enhancing coded computing with paradigm-specific adversary-resistance mechanisms. In contrast to prior work in this field, which typically addresses privacy, robustness, or decentralization in isolation, our approach provides a common framework that explicitly faces privacy leakage and malicious behavior under realistic attack scenarios.

\subsection{Privacy-Enhancing Technologies for Distributed Learning}
\label{sec:related-privacy}

A substantial body of work has explored the use of cryptographic Privacy-Enhancing 
Technologies to protect distributed learning. Homomorphic encryption enables training 
over encrypted data or model updates, providing strong theoretical privacy 
guarantees~\cite{bonawitz2016practical, aono2017privacy}. The application of these 
techniques to federated learning architectures has been recently analyzed in the 
literature~\cite{xie2024efficiency}, considering approaches based on either single 
keys~\cite{aziz2023exploring} or multiple keys~\cite{ma2022privacy}. 
Secure multi-party computation has similarly been applied to federated learning to prevent
information leakage during the aggregation process~\cite{bonawitz2016practical, tran2023novel,
mou2021verifiable}. Some of these approaches have been successfully applied across different 
application domains, including finance~\cite{byrd2020differentially}, mobile 
networks~\cite{mugunthan2019smpai}, and IoT-enabled manufacturing~\cite{kanagavelu2020two}. 
However, these approaches often incur significant computational and communication overheads, 
which limit scalability and practical deployment, particularly for large models or resource-
constrained environments. 

Differential privacy has also been widely adopted to mitigate information leakage in federated learning. In particular, several approaches perturb model updates or gradients to provide formal privacy guarantees~\cite{wei2020federated}. These techniques have been applied across a wide range of application domains, including the Internet of Vehicles (IoV)~\cite{zhao2020local} and personalized services~\cite{hu2020personalized}. While differential privacy provides formal guarantees, it introduces a privacy--utility trade-off and does not inherently
address malicious manipulation or Byzantine behavior. Moreover, privacy guarantees typically degrade under repeated training rounds or adaptive adversaries~\cite{wei2020federated}.

In contrast to the aforementioned cryptographic- or noise-based approaches, coded computing techniques introduce structural obfuscation through information mixing and approximation. These methods provide privacy-enhancing properties
without relying on heavy cryptographic primitives, making them attractive for deployable systems. Our work builds on this line by integrating GPBACC as a model-agnostic PET and validating its privacy properties empirically under explicit attacks.

\subsection{Robust Aggregation in Federated Learning}
\label{sec:related-aggregation}

Federated learning is particularly vulnerable to poisoning and Byzantine attacks due to its reliance on distributed client updates. Robust aggregation techniques such as Krum, Multi-Krum, Trimmed Mean, and coordinate-wise median have been proposed to mitigate the impact of malicious participants~\cite{blanchard2017machine, yin2018byzantine}. Subsequent work has explored adaptive and statistically robust variants, as well as defenses against backdoor attacks~\cite{bagdasaryan2020backdoor, Pillutla2022}. Some initiatives perform well under IID settings, but their performance degrades in non-IID scenarios~\cite{li2023experimental}. Other approaches have been tailored for secure hardware architectures, such as Intel SGX~\cite{zhao2021sear}, or have been specifically designed for industrial environments~\cite{li2021byzantine}.

While these methods claims to improve robustness, they are typically designed independently of privacy considerations and assume access to raw or lightly processed model updates. As a result, they remain susceptible to privacy
inference attacks and are tightly coupled to specific aggregation settings. In this work, we do not propose new robust aggregation rules; instead, we combine existing robust aggregation strategies with GPBACC to jointly address privacy leakage and malicious behavior in federated learning.

\subsection{Adversary Resistance in Decentralized Learning}
\label{sec:related-adversary}

Decentralized learning, where there is not a master element that owns the data, removes the central aggregation point and, consequently, complicates adversary detection and verification. Prior work on decentralized gossip-based learning
has focused primarily on convergence and scalability, often under benign assumptions~\cite{lian2017can}. Byzantine-resilient decentralized learning has received comparatively less attention, and existing solutions~\cite{yang2019adversary,yang2020adversary} frequently rely on strong assumptions about network connectivity or honest majorities. 

Verification-based approaches, including consistency checks and redundancy, have been explored to detect faulty or malicious participants in distributed systems~\cite{xing2025zero}. Group testing techniques have been applied in other domains to efficiently identify adversarial actors with limited overhead~\cite{du2013group}. Our work adapts these ideas to decentralized learning by combining approximate decode-and-compare with group testing, enabled
by the coded structure introduced through GPBACC. This allows adversary detection and isolation without requiring full decoding or a trusted coordinator.

\subsection{Coded Computing for Machine Learning}
\label{sec:related-coded}

Coded computing has been extensively studied as a means to mitigate stragglers and failures in distributed computation~\cite{lee2017speeding,li2015coded, Wu2025}, while some are able to ensure privacy guarantees~\cite{LCC:YLRK19, ALCC:SMA21}. Some work has applied coding techniques directly to machine learning workloads, including gradient coding~\cite{tandon2017gradient}. These approaches primarily target efficiency and fault tolerance, with privacy considered as a secondary or implicit benefit, and only work with specific classes of functions, which reduces its usability in common ML.

Berrut-based rational approximations have been explored for their numerical stability and efficiency in interpolation tasks~\cite{berrut2004barycentric}, being BACC~\cite{BACC:NA20} the most notorious scheme, capable of distributing the computation of arbitrary functions while resisting stragglers. Our previous research work, Generalized Privacy-aware Berrut Approximated Coded Computing (GPBACC)~\cite{gpbacc25}, was built on these foundations to enable generalized approximate coded computation suitable for machine learning. Unlike prior coded computing approaches, we emphasize GPBACC as a privacy-enhancing primitive and demonstrates how it can be combined with adversary-resistance mechanisms across both federated and decentralized learning paradigms.

\section{Preliminaries}
\label{sec:prel}
\subsection{Generalized Privacy-aware Berrut Approximated Coded Computing (GPBACC)}
\label{sec:bacc}

The Berrut Approximated Coded Computing (BACC) scheme~\cite{BACC:NA20} enables 
the distributed computation  of an arbitrary objective function 
$f : \mathbb{V} \rightarrow \mathbb{U}$ over input data $\mathbf{X} = 
(X_0, \dots, X_{K-1})$,  where $\mathbb{U}$ and $\mathbb{V}$ are vector spaces of 
real matrices. The scheme operates on a network composed of one master node (data 
owner) who commissions $N$ worker nodes the task of computing $f(\mathbf{X}) = 
\bigl( f(X_0), \dots, f(X_{K-1}) \bigr)$. Specifically, BACC approximates
\begin{equation*}
    \tilde{f}(\mathbf{X}) \approx \bigl(f(X_0), f(X_1), \dots, f(X_{K-1})\bigr)
\end{equation*}
in each component with bounded numerical error, offering stability even for large 
numbers of workers. It tolerates stragglers, as the approximation error depends on the 
number of received results, and can represent not only polynomial functions but 
also other arbitrary functions $f$ under mild regularity  assumptions too~\cite{Qiu2025}. 
The protocol proceeds through three phases, as shown in Figure~\ref{fig:pbacc}:

\begin{figure}
    \centering
    \includegraphics[width=0.99\columnwidth]{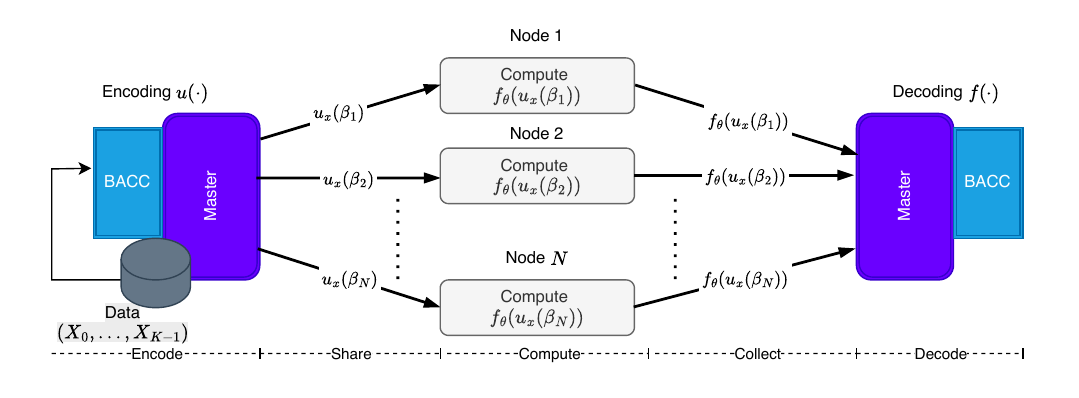}
    \caption{Phases of the BACC scheme.}
    \label{fig:pbacc}
\end{figure}

\paragraph{Encoding and Sharing}
The master node encodes the data vector $\mathbf{X}$ using a rational interpolation 
function based on the Berrut barycentric formulation. Given a vector 
$\boldsymbol{\alpha} = (\alpha_0, \dots, \alpha_{K-1})$ of distinct interpolation points,
define the rational functions
\begin{equation*}
    \tilde{q}(\alpha_i, z) := \frac{(-1)^i}{z - \alpha_i}, \qquad Q(\boldsymbol\alpha, z) := \sum_{i = 0}^{K - 1} \tilde{q}(\alpha_i, z),
\end{equation*}
and let $q(\alpha_i, z) = \tilde{q}(\alpha_i, z) / Q(z\boldsymbol, )$. The Berrut rational interpolation is
\begin{equation}
\label{eq:bacc-encoding}
 u(z) =  \sum_{i=0}^{K-1} q(\alpha_i, z) \, X_i,
\end{equation}
where $\boldsymbol\alpha$ are selected as the Chebyshev points of the first kind
$\alpha_j = \operatorname{Re}\bigl( \exp{((2j + 1) \pi / (2K))\bigr)}$, $j = 0,\dots,K-1$.
It follows by direct substitution that $u(\alpha_j) = X_j$, $\forall j$.
Then, $N$ encoder mapping points $\boldsymbol{\beta} = \{\beta_0, \dots, \beta_{N-1}\}$ 
are chosen equal to the Chebyshev points of the second kind:
$\beta_j = \operatorname{Re}\bigl( \exp{j \pi / (N - 1)}\bigr)$, $j = 0,\dots,N-1$.
The master evaluates $u(\beta_j)$ for each $\beta_j$ and sends this encoded share to 
worker $j$.

\paragraph{Computation and Communication}
Each worker $j$ computes the result $v_j = f\big(u(\beta_j)\big)$ and sends it back 
to the master node.

\paragraph{Decoding}
Upon receiving the first $n$ responses from the nodes $\mathcal{F} \subset [N]$,
$|\mathcal{F}| = n$, the master node reconstructs the approximated outputs through 
the Berrut rational interpolation
\begin{equation}
  \label{eq:bacc-decoding}
  r_{\mathrm{Berrut}, \mathcal{F}}(z) =  \sum_{i=0}^{n} q(\beta_i, z) \, 
  f\big(u(\tilde{\beta}_i)\big),
\end{equation}
where $\tilde{\beta}_i \in \mathcal{S} = \{\cos\frac{j\pi}{N-1}, j \in \mathcal{F}\}$.
The master estimates $f(X_i) \approx r_{\mathrm{Berrut}, \mathcal{F}}(\alpha_i)$, for 
all $i \in [0, K-1]$.

A weakness of BACC for FL and DL workflows is that any node able to observe the 
encoded shares can infer information about $\mathbf{X}$, that is, no 
privacy guarantee exists. The Privacy-Aware extension of BACC solves this 
problem with the introduction of Gaussian random terms into the encoding process.
In PBACC the encoding function is given by
\begin{equation}
  \label{eq:pbacc-encoding}
  u(z)
     = \sum_{i = 0}^{K+T-1} q(\alpha_i, z) W_i,
\end{equation}
where $W_i = X_i$ for $i = 0, \dots, K-1$ and $W_i = R_{i - K}$ for $i = K, \dots, T - 1$ 
and $\{R_i\}$ are iid random samples drawn from $\mathcal{N}(0, \sigma_n^2/T)$. The 
data interpolation  points $\{\alpha_0,\dots,\alpha_{K-1}\}$ are as in GBACC,  while the 
noise points  $\{\alpha_K, \dots, \alpha_{K+T-1}\}$ are shifted Chebyshev points
$\alpha_{K+j} = b +  \operatorname{Re}\bigl( \exp{(2j+1) \pi / (2T) }\bigr)$, 
for $j = 0,\dots,T-1$, where $b \in \mathbb{R}$ defines the shift.

Neither Berrut coded computing nor its private version PBACC support  multi-source
configurations and tensor data, and cannot be used directly in many machine learning
computing tasks. \emph{Generalized PBACC} (GPBACC) extends the encoding mapping so 
that its range accepts arbitrary-rank inputs. Consider $N$ nodes, each in possession
of a private data tensor $\mathbf{X}^{(i)}_{k_0 k_1 \dots k_{L-1}} \in 
\mathbb{R}^{K \times k_1 \times \dots \times k_{L-1}}$, where $i$ indexes the data 
owner and $L$ denotes the tensor rank. In GPBACC, the encoding operation for node $i$ is
performed as
\begin{equation}
  \label{eq:gpbacc-encoding}
  \begin{aligned}
  u_{\mathbf{X}^{(i)}}(z)  &= \sum_{j = 0}^{K - 1} q(\alpha_j, z) \, X^{(i)}_{j k_1 
     \ldots k_{L-1}} \\ &+ \sum_{j = 0}^{T - 1} q(\alpha_{K + j}, z) \, R^{(i)}_{j k_1 \ldots k_{L-1}},
  \end{aligned}
\end{equation}
where $\alpha_j$ are the interpolation points as before,  and $R^{(i)}_{j k_1 \ldots 
k_{L-1}} \sim \mathcal{N}(0, \sigma_n^2/T)$ are random tensors. The function
$u_{\mathbf{X}^{(i)}}(z)$ is evaluated at $\beta_j = \cos\!\left(\frac{j\pi}{N-1}\right)$,
the Chebyshev points of the second kind, producing shares 
$u_{\mathbf{X}^{(i)}}(\beta_j)$ distributed to all nodes. Each node $i$ computes
$f\big(u_{\mathbf{X}^{(0)}}(\beta_i), u_{\mathbf{X}^{(1)}}(\beta_i), 
\dots, u_{\mathbf{X}^{(N-1)}}(\beta_i)\big)$ and sends the result to the master 
node, which reconstructs the outputs using the Berrut decoder
\begin{equation}
  \label{eq:gpbacc-decoding}
  r_{\mathrm{Berrut}, \mathcal{F}}(z) = \sum_{i = 0}^{n} q(\beta_i, z) \, 
  f\big( u_{\mathbf{X}^{(0)}}(\tilde{\beta}_i), 
  \dots, u_{\mathbf{X}^{(N-1)}}(\tilde{\beta}_i) \big),
\end{equation}
where $\tilde{\beta}_i \in \mathcal{S} = \{\cos\frac{j\pi}{N-1}, j \in \mathcal{F}\}$ 
are the evaluation points of the fastest nodes.
Finally, the master approximates
\begin{equation}
  f\big(X^{(0)}_{j k_1 \ldots k_{L-1}}, \dots, X^{(N-1)}_{j k_1 \ldots k_{L-1}}\big)
  \approx r_{\mathrm{Berrut}, \mathcal{F}}(\alpha_j),
\end{equation}
for all $j \in [0, K-1]$.

\subsection{Privacy metric and Privacy Guarantees}

We adopt an  information–theoretic leakage metric based on the worst-case achievable mutual 
information for a subset of $c$ colluding semi-honest nodes. A semi-honest node is
any participant in the system that follows the rules of the computing protocol but
it is interested in obtaining the original input $\mathbf{X}$ by processing the
messages exchanged during the execution of the distributed computation. 

To derive the information-theoretic bounds, we recall that the encoding process is
analogous to an equivalent Multi-Input Multi-Output (MIMO) channel with $K$ 
transmitter antennas (the encoded input components) and $c$ receiver antennas 
(the colluding nodes)~\cite{MIMOCapacity:SPM02}. Following this formulation, the 
injected Gaussian noise of GPBACC acts as additive channel noise that limits the
information capacity of the system. Assuming that the encoded noise is independent 
and identically distributed according to $\mathcal{N}(0,\sigma_n^2/T)$,
the system can be viewed as an Additive White Gaussian Noise (AWGN) vector channel.

The capacity $C$ of this MIMO channel is by definition an upper-bound on the 
mutual information between the transmitted signal $\mathbf{X}$ and the received signal 
$\mathbf{Y}$, and it is given by
\begin{equation}
  \label{eq:mimo-capacity}
  C = \sup_{P_{\mathbf{X}}} I(\mathbf{Y}; \mathbf{X}) = \log_2 
  \bigl| I_c + P H \Sigma_{\mathbf{Z}}^{-1} H^{\dagger} \bigr|,
\end{equation}
where $H$ denotes the encoding matrix, $P$ the maximum transmission power per input,
$\Sigma_{\mathbf{Z}}$ the noise covariance matrix, and $I_c$ the $c \times c$ identity
matrix. The determinant $|\cdot|$ expresses the total channel gain observed by the 
colluding subset.

Leveraging this framework, we define the worst-case achievable mutual 
information $I_L$ as the maximum information attainable by any set $\mathcal{C} \subset [N]$
colluding workers:
\begin{equation}
  I_L \triangleq  \max_{\mathcal{C}} \sup_{P_{\mathbf{X}}:\,\|X_i\|\le s}
  I(\mathbf{Y}_{\mathcal{C}}; \mathbf{X}),
\end{equation}
where $s$ bounds the magnitude of the input signal. By the power constraint 
$\mathbb{E}[\|X_i\|^2] \le s^2$, we can write
\begin{equation}
  \label{eq:IL-bound}
  I_L \le \max_{\mathcal{C}} \sup_{P_{\mathbf{X}}:\,\mathbb{E}[\|X_i\|^2]\le s^2}
  I(\mathbf{Y}_{\mathcal{C}}; \mathbf{X}).
\end{equation}
Combining~\eqref{eq:mimo-capacity} and~\eqref{eq:IL-bound}, the upper bound on the information leakage becomes
\begin{equation}
  \label{eq:IL-final}
  I_L \le \max_{\mathcal{C}} \log_2 \Bigl| I_c + \frac{s^2 T}{\sigma_n^2}
  \tilde{\Sigma}_{\mathcal{C}}^{-1} \Sigma_{\mathcal{C}} \Bigr|,
\end{equation}
where $\Sigma_{\mathcal{C}}$ and $\tilde{\Sigma}_{\mathcal{C}}$
are the covariance matrices corresponding to the encoded data and noise, respectively:
\begin{equation}
  \Sigma_{\mathcal{C}} \triangleq \begin{pmatrix}
  q_{0}(\beta_{i_1}) & \cdots & q_{K-1}(\beta_{i_1}) \\ \vdots & \ddots & \vdots\\
  q_{0}(\beta_{i_c}) & \cdots & q_{K-1}(\beta_{i_c})
  \end{pmatrix},
\end{equation}
\begin{equation}
  \tilde{\Sigma}_{\mathcal{C}} \triangleq \begin{pmatrix}
  q_{K}(\beta_{i_1}) & \cdots & q_{K+T-1}(\beta_{i_1})\\
  \vdots & \ddots & \vdots\\
  q_{K}(\beta_{i_c}) & \cdots & q_{K+T-1}(\beta_{i_c})
 \end{pmatrix},
\end{equation}
and $\{\beta_{i_h}\}$ being the evaluation points of the colluding nodes.

Finally, the privacy metric adopted in this paper is the information leakage normalized
per data element
\begin{equation}
  \label{eq:il-normalized}
  i_L = \frac{I_L}{K}.
\end{equation}
A GPBACC configuration is considered $\epsilon$-secure if $i_L \le \epsilon$ for the 
given parameters $(c, T, \sigma_n)$. This analytical bound it will be 
empirically validated in Section~\ref{sec:results} through the evaluation against
robust Membership Inference Attacks.

\subsection{Verifiable Distributed Computing}

\subsubsection{Decode-and-Compare (DC)}
\label{sec:decode-compare}

The Decode-and-Compare strategy is a classical verification mechanism
for distributed coded computing schemes that produces exact reconstructions.
It enables the master node to validate the correctness of received intermediate results
without requiring cryptographic proofs or redundant recomputation.

Let $I = \{ I_1, I_2, \dots, I_{m} \}$ denote the set of intermediate results
received by the master node from $m$ workers, where each result $I_i$
corresponds to the encoded evaluation $f(u(\beta_i))$. Within $I$, a subset $I^v$ 
contains the correct results, and a subset $I^x$ the incorrect (erroneous or adversarial)
results, so $I = I^v \cup I^x$, and $I^v \cap I^x = \emptyset$. Then, the verification 
principle for DC works as follows. If the coded computing scheme supports exact 
decoding from any subset of $k$ valid intermediate results, then for $m > k$ the 
master can form multiple verification sets $S \subseteq \{1, \dots, m\}$ 
of size $|S| = m$. For each such set $S$, the master reconstructs the final result 
using the generic reconstruction operator 
\begin{equation}
    r_S(z) = \mathsf{Decode}\bigl( \{ I_i : i \in S \} \bigr),
\end{equation}
where $\mathsf{Decode}(\cdot)$ represents the interpolation or decoding process
of the underlying coded computing scheme.

The DC strategy relies on the deterministic nature of $r_S(\cdot)$ to
verify the consistency of the computation, since if the decoding operation is 
exact and two distinct verification sets $S_1$ and $S_2$ consist only of 
correct intermediate results, then their independent reconstructions must coincide
$r_{S_1}(z) = r_{S_2}(z)$, for every $z \in \mathcal{Z}$,
where $\mathcal{Z}$ is the domain of evaluation (typically the set of decoder points).
Hence, in DC the master node tests whether any two reconstructed results match
\begin{equation}
  \label{eq:dc-match}
 \exists \, S_1, S_2 \subseteq \{1, \dots, m\}, \; |S_1| = |S_2| = m : 
  r_{S_1}(z) = r_{S_2}(z) 
\end{equation}
for all $z \in \mathcal{Z}$. As soon as two identical reconstructions are found,  
the corresponding decoded value is accepted as the correct result, and the decoding 
process terminates.

Therefore, under the assumptions that: (i) the decoding function 
$\mathsf{Decode}(\cdot)$ yields exact results  for any $k$ correct inputs; 
(ii) distinct verification sets containing at least one incorrect result
produce distinct decoded outputs, DC ensures deterministic detection of 
correctness without false acceptance.

Nevertheless, in approximate or noisy schemes—such as GPBACC, which
introduce controlled perturbations to achieve privacy, equality 
in~\eqref{eq:dc-match} may not hold even for honest nodes. This motivates 
the development of an Approximate Decode-and-Compare (ADC) variant (see
Sect.~\ref{sec:adc-gt-design}), which introduces tolerance intervals based on the decoding 
error envelope and confidence thresholds. 

\subsubsection{Group Testing (GT)}
\label{sec:group-testing}

Since schemes like GPBACC achieve higher accuracy reconstruction as the number of 
correct intermediate results increases, it is advantageous to precisely identify and 
exclude malicious nodes rather than merely detecting inconsistencies. To this end, 
the Approximate Decode-and-Compare mechanism may be complemented with a Group 
Testing strategy, for efficient localization of adversarial workers whose outputs should 
be disregarded during decoding.

A GT protocol is an efficient algorithm to identify defective items 
within a large population through a minimal number of collective tests. Instead of 
examining each item individually, GT performs tests over groups (pools) of items and infers 
the defective set based on the pattern of positive and negative outcomes. 

Let $\mathcal{N} = \{1, 2, \dots, N\}$ denote the set of all workers, 
among which an unknown subset $\mathcal{A} \subseteq \mathcal{N}$ of size $s \ll N$
is adversarial (or faulty). A pooling matrix $\mathbf{G} \in \{0,1\}^{M \times N}$ 
defines which workers participate in each of the $M$ tests, where $G_{p,i} = 1$ if 
worker $i$ is included in pool $p$ and $0$ otherwise. For each pool $p$, a collective 
test is performed, and its binary outcome $y_p \in \{0,1\}$ indicates whether at least 
one adversarial element was present in the pool:
\begin{equation}
  y_p = \begin{cases}
   1, & \text{if } \exists i \in \mathcal{A} \text{ with } G_{p,i}=1,\\[2pt]
   0, & \text{otherwise.}
 \end{cases}
\end{equation}
The goal is to reconstruct the unknown set $\mathcal{A}$  from the observation 
vector $\mathbf{y} = (y_1, \dots, y_M)$.

Classical GT designs rely on combinatorial properties such as $d$-disjunct matrices, 
which guarantee that up to $s$ adversarial workers can be uniquely 
identified if $M = O(s \log N)$ tests are performed. Alternatively, random Bernoulli
designs with $G_{p,i} \sim \mathrm{Ber}(\theta)$ achieve comparable performance with 
high probability, offering scalability and simplicity in implementation. Decoding 
is then performed using logical rules like Combinatorial Orthogonal Matching Pursuit
(COMP), probabilistic inference (Definite Defectives, DD), or optimization-based
formulations (e.g., sparse recovery).

In the context of distributed computing, each pool corresponds to a subset of 
intermediate results combined and decoded together using the reconstruction 
function $r_S(\cdot)$. If the reconstruction is inconsistent with expected behavior,
the corresponding pool is labeled as failed. By analyzing the pattern of failing and
passing pools through GT decoding, the master node can efficiently localize the 
malicious workers. Error-resilient GT formulations are particularly important in
approximate-coded schemes (such as GPBACC), where numerical noise may induce false 
alarms or missed detections. Hence, integrating GT with Decode-and-Compare verification
provides a scalable and fault-tolerant mechanism to detect and isolate adversarial 
nodes.

\section{Threat Model}
\label{sec:threat-model}

The distributed GPBACC scheme may be deployed under a Federated Learning
system, where workers possess private data and exchange model updates with the master node, 
and in Decentralized Learning systems, where each node processes shared data owned by 
a master. For both cases, we pose a threat model wherein adversaries may attempt to 
compromise the privacy or the integrity of the computation.
We distinguish two classes of adversaries according to their objectives and accessible 
information:
\begin{itemize}
\item Semi-honest (honest-but-curious) adversaries, who correctly follow the GPBACC 
protocol but attempt to infer information about other nodes' or master's private data.
Semi-honest participants may collude, sharing their encoded inputs or outputs.  
The privacy requirement of GPBACC is to ensure that the information leakage per data 
point, $i_L$, remains bounded by a target threshold $\epsilon$ for any subset $\mathcal{C}
\subseteq \mathcal{N}$ of $|\mathcal{C}| = c$ colluding nodes.

\item Malicious (Byzantine) adversaries, those who actively deviate from the protocol 
by manipulating their computations, sending incorrect, noisy, or adversarially crafted 
intermediate results. Such nodes may attempt to degrade the accuracy of the global model 
or output (poisoning attacks). They could also target the decoding to fail by returning 
inconsistent or divergent values, or remain stealthy by mimicking honest outputs within a small 
perturbation margin.
\end{itemize}

The two types of adversaries pursue privacy or integrity attacks, respectively. 
Semi-honest adversaries aim to reconstruct or infer private information about
the encoded data $\mathbf{X}$, local updates, or sensitive model parameters. In Federated
Learning, this corresponds to classic Membership Inference Attacks (MIA) or Gradient 
Leakage Attacks, where adversaries use the received model updates to infer whether a 
specific data point was part of the training set. GPBACC’s privacy objective is to prevent 
the success of such inference through noise-induced encoding. Malicious adversaries seek 
to compromise the correctness of the aggregated result,  e.g., by injecting corrupted 
gradients or flipped labels in FL, or by sending arbitrary outputs in DL.
The adversaries are assumed to have the following knowledge and operational limitations:
\begin{itemize}
\item Semi-honest adversaries may share their local encoded data or intermediate results, 
but cannot access the raw data or randomness of other nodes.
\item Malicious adversaries may craft arbitrary outputs, but cannot compromise the master 
node or the secure communication channels.
\end{itemize}
Furthermore, the encoding and decoding algorithms, interpolation points, and the 
public parameters of the scheme are publicly know, but the random coefficients $R_i$ used 
in the encoding and the private inputs of other participants are not observable by other 
nodes.
 
The design goal of GPBACC is to guarantee that $i_L \leq \epsilon$ in presence of up to $c$ 
colluding semi-honest workers, and that up to $b$ malicious adversaries are resisted exactly 
as if they were stragglers, so adversarial nodes have the less possible impact on the 
precision of the reconstruction.

\begin{figure*}[t]
    \centering
    \includegraphics[width=0.9\linewidth]{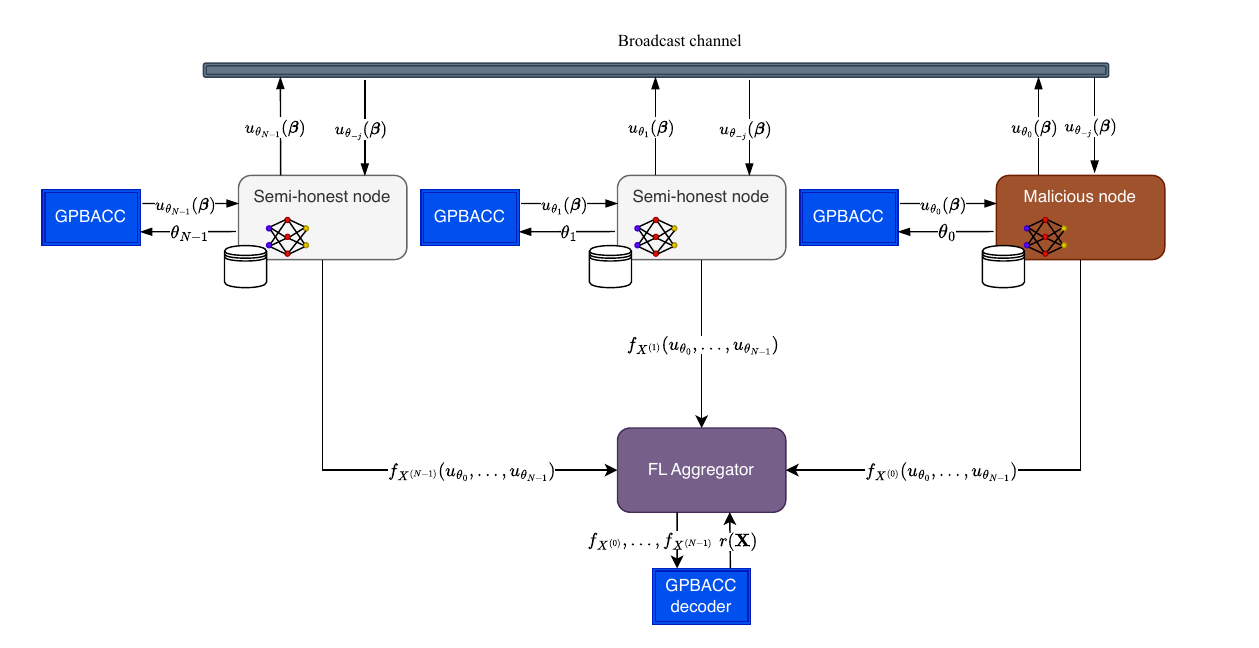}
    \caption{GPBACC over Federated Learning}
    \label{fig:gpbacc_over_fl}
\end{figure*}

\section{Privacy Preservation and Byzantine Resilience in GPBACC}
\label{sec:gpbacc-adversary-resistance}

GPBACC supports adversary-resilience under federated and distributed learning alike. Though
the same privacy-preserving coded computing are leveraged as common technical basis, 
the mechanisms required to guarantee integrity differ substantially. This Section describes 
our proposed solution for practical privacy and robust, verifiable computation with GPBACC.

\subsection{Private aggregation in Federated Learning}

In FL, our encoded scheme integrates naturally with robust aggregation strategies, just by 
embedding the aggregation rule (e.g., median, trimmed mean, Krum, Multi-Krum) directly 
as the target function~$f$ evaluated by the workers. We show below that setting the coding 
compression dimension to $K = 1$ enables a clean per-coordinate decoding structure while 
still allowing the master to reconstruct the aggregated update with high numerical precision. 
Under this condition, robust aggregation mitigates possible poisoning attacks, while 
GPBACC simultaneously provides input privacy and coded redundancy. The end-to-end
workflow for this setting is illustrated in Fig.~\ref{fig:gpbacc_over_fl} for a simple 
setup with three clients (two honest, one adversarial) and a central aggregator.

The computational procedure for private aggregation comprises the following steps.
Assume that all data owners have received an initial global model from the aggregator, and 
have performed the local training using their local data. At this point, each data owner $i$ 
encodes its model weights $\theta^{(i)}$, after receiving a set of $N$ shares 
$u_{\theta^{(i)}} = [u_{\theta^{(i)}}(\beta_0), \dots,  u_{\theta^{(i)}}(\beta_{N-1})]$.
Without loss of generality, let data owner $0$ be an adversary, its model weights 
$\theta'^{(0)}$ deliberately altered by some poisoning strategy and its corresponding
encoded shares computed as $u_{\theta'^{(0)}} = [u_{\theta'^{(0)}}(\beta_0), \dots, 
u_{\theta'^{(0)}} (\beta_{N-1})]$.  In order to keep the coded elements decodable and to
preserve a simple per-coordinate decoding interpretation, the coding compression dimension 
is set  to $K=1$. Hence, each owner's encoded rational function compresses the first 
dimension to a single interpolation element (no multi-slicing along the first index). 
This is done as follows:
\begin{itemize}
\item[(i)] The encoded share $u_{\theta^{(i)}}(\beta_j)$ represents a single coded object 
whose entries correspond to coordinates of $\theta^{(i)}$. The Berrut decoder is 
then applied coordinate-wise (or on flattened tensors).
\item[(ii)] With $K=1$, any decoding set $S$ of size $k$ corresponds to exactly thhe minimal 
number of points required to recover the coded element,
which avoids inter-block interference.
\end{itemize}
This choice keeps the structure of GPBACC compatible with standard aggregation hence
it allows workers to compute robust aggregators on the received encoded vectors
coordinate-wise (or via per-coordinate operations) and permits the master (aggregator)
to apply Berrut decoding coordinate-wise to obtain the aggregated coordinates.

In the communication phase, each data owner $i$ exchanges $u_{\theta^{(i)}}(\beta_j)$ with 
every other data owner $j \neq i$. Then, the computation phase starts whereby each data 
owner $j$ uses its corresponding shares $[u_{\theta^{(0)}}(\beta_j), \dots, u_{\theta'^{(N-1)}}
(\beta_j)]$ for computing the target function value $v_j = f(u_{\theta^{(0)}}(\beta_j), \dots,
u_{\theta'^{(N-1)}}(\beta_j)) $, which in this case will be some robust aggregation strategy.
We consider two representative families of aggregations compatible with GPBACC.
\begin{enumerate}
\item \emph{Coordinate-wise robust rules (Median, Trimmed-Mean)}. These aggregators 
operate independently per coordinate:
\begin{equation*}
  \operatorname{Agg}_{\text{coord}}(\{ \theta^{(i)} \}_{i =0}^{N - 1})_r = \operatorname{stat}
  \bigl(\{ \theta^{(i)}_r \}_{i = 0}^{N - 1}\bigr), r=1, \dots,d,
\end{equation*}
where $\operatorname{stat}(\cdot)$ is a simple estimator, like median, trimmed mean, etc.
    
\item \emph{Selection-based rules (Krum, Multi-Krum)}. Krum 's robust aggregation 
algorithm selects one client update that is most representative (minimizes summed
distances to nearest neighbors); instead, multi-Krum averages several such selections. 
Formally, Krum's algorithm returns
\begin{equation*}
\operatorname{Krum}(\{ \theta^{(i)} \}) = \theta^{(j^*)}, j^* = 
\arg\min_j \sum_{m = 1}^{N - b - 2} \|\theta^{(j)} - \theta^{(m)}\|^2,
\end{equation*}
with appropriate neighbor selection.
\end{enumerate}
Once the computation in the encoded domain is finished, each data owner will send 
its locally computed result $v_j$ to the Aggregator.
Lastly, the aggregator approximately reconstructs the value of $f(\theta^{(0)}, \dots, 
\theta^{(N-1)})$ with the decoding function. 

\emph{Remarks}: Robust aggregators increase per-worker compute costs, since each 
worker must evaluate the aggregator on the set of encoded inputs it receives. Also,
it is worth noting that some robust selection rules (e.g., Krum) are sensitive to the
decoding noise, because selection depends on pairwise distances between the received 
results, unlike coordinate-wise aggregators (i.e., median, trimmed mean), that tolerate 
larger coordinate-wise decoding variance. Increasing the noise parameter $\sigma_n$ (for 
privacy) or increasing redundancy $T$ reduces decoding precision and can affect Krum's
aggregation more severely than coordinate-wise methods.

\subsection{Adversary-resistance with Approximate Decode-and-Compare \& Group Testing}
\label{sec:adc-gt-design}

\begin{figure*}[t]
    \centering
    \includegraphics[width=0.99\textwidth]{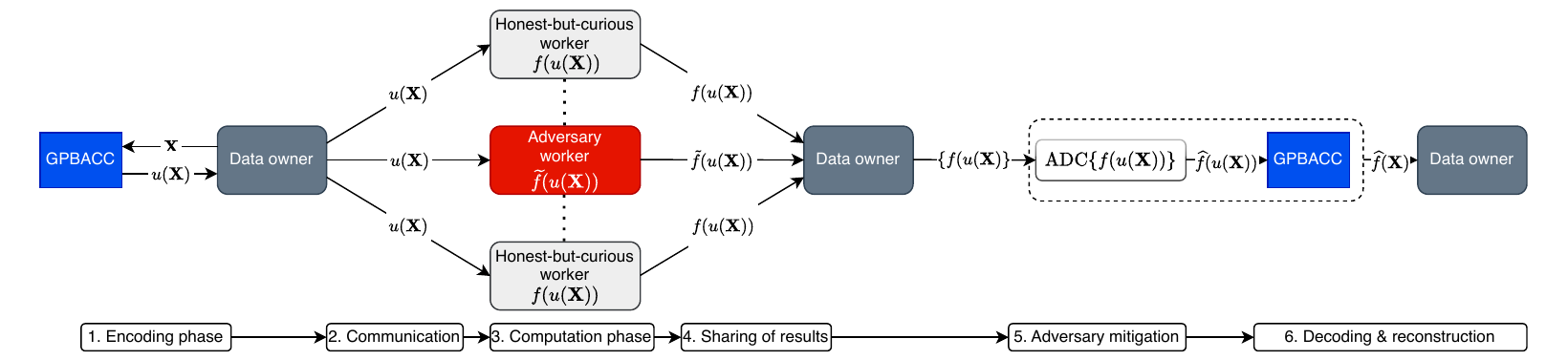}
    \caption{GPBACC over Distributed Learning}
    \label{fig:gpbacc_over_dl}
\end{figure*}

In DL, there is a single data owner that shares data, so no aggregator entity is
involved. Instead, correctness must be achieved through verification mechanisms 
applied directly to coded intermediate results exchanged from the workers. For this setting, 
we introduce a dedicated verification pipeline based on ADC combined with GT. ADC provides 
a principled consistency test adapted to GPBACC’s approximate reconstruction error, while GT
efficiently localizes adversarial workers by pooling subsets of worker outputs. Together, 
ADC and GT enable a prune-and-refine decoding procedure that removes corrupted
contributions and reconstructs the output using only verified inputs. The full DL workflow 
is depicted in Fig.~\ref{fig:gpbacc_over_dl} for a configuration with three workers and one 
data owner (master).

The proposed scheme operates according to this workflow. For the encoding phase, consider
a initial situation where the master data owner wants to exchange their data to a set 
of workers to offload the training of a ML model. At this point, the data owner encodes 
the dataset $X$, generating a set of $N$ shares $u_{X} = [u_{X}(\beta_0), \dots, 
u_{X}(\beta_{N-1})]$, for all $i \in [ N ]$. Differently from a centralized aggregation, 
$K$ can now be tuned for balancing the communication cost versus the numerical precision
of the approximate computations.

Next, the data owner transmits each share $u_{X}(\beta_i)$ to worker $i$, who uses its
corresponding share $u_{X}(\beta_i)$ for computing the target function 
$v_i = f(u_{X}(\beta_i)) = v_i$, which in our framework will be some updating rule
for a local model (e.g., regressions, CNNs, VAEs, etc; the specific choice of the local 
model is independent on the encoded computations). Assuming worker $0$ is malicious,
its corrupted result will be denoted as $v'_0$. Following the scheme, each worker $i$ 
will exchange its local computed result $v_i$ with the master data owner,
and it will be the master's task to detect and filter out the Byzantine participants' results
with the proposed GT+ADC prune-and-refine strategy for obtaining an approximate result 
$f(X)$ contributed only by honest participants.

More specifically, the proposed ADC+GT computing pipeline detects inconsistent worker 
outputs in approximate-coded scenarios, localizes malicious workers using pooling, and 
prunes them to re-decode with a cleaner set. The design leverages the decoding error 
envelope of Berrut interpolation to derive principled thresholds. To this end, let 
$r_S(\cdot)$ denote the generic reconstruction operator obtained from any verification 
subset $S$ in our approximate Decode-and-Compare. For a reconstruction
$\widehat{\mathbf{f}}_S := r_S(\alpha)$ evaluated at a decoder point $\alpha$, denote by
$E(z,s,N,f)$ the worst-case decoding error envelope when decoding in the presence of $s$
stragglers  (equivalently, using a subset of size $|S| = N - s$), interpolation 
geometry parameterized by $z$, and target function $f$. Formally, assume that 
$\| \widehat{\mathbf{f}}_S - \mathbf{f}^\star\| \le E(z,s,N,f)$,
where $\mathbf{f}^\star$ is the ideal (noiseless) aggregate. Given two verification
subsets $A$ and $B$ of identical size, by the triangle inequality
$ \|\widehat{\mathbf{f}}_A - \widehat{\mathbf{f}}_B\| \le \|\widehat{\mathbf{f}}_A -
\mathbf{f}^\star\| + \|\mathbf{f}^\star - \widehat{\mathbf{f}}_B\| \le 2 E(z,s,N,f)$, 
 we could adopt the detection threshold
\begin{equation}
  \label{eq:adc-threshold}
  \tau_{g} \;=\; 2\,E(z,s,N,f)
\end{equation}
and declare two reconstructions approximately equal when their distance is
$\le \tau_{g}$. Clearly, the bound $\tau_{g}$ is a worst-case scenario for the decoding 
error, therefore it will be very loose in many real scenarios, thus misleading the 
strategy toward false negatives. For this reason, in this work we use a bootstrap strategy 
to improve the tightness of the ADC acceptance rule:
\begin{enumerate}
\item The master randomly subtracts a subset $\hat{u}_{X}$ of length $d$ from the $N$ 
shares of the encoded dataset $u_{X}$, iterating over the encoded dimension. This means that, 
if the original dataset is a tensor of size $k_0\times k_1\times \dots \times k_{L-1}$, each
encoded share will have a compressed size of $K\times k_1\times \dots \times k_{L-1}$, and 
the reduced subset will have shares of size $d\times k_1 \times \dots \times k_{L-1}$.  
Note that $d \leq K$, and as $d$ increases, the threshold calculated will be more accurate 
in exchange for bootstrap time.

\item The master computes $d$ model executions (using the initial model weights) over the
$N$ subtracted shares, and for each execution, divides the results in $p$ groups of length 
$d$.

\item For each model execution, the master decodes each of the groups, obtaining the group 
of results $\mathcal{V} = [\mathcal{V}_0, \mathcal{V}_1, \dots, \mathcal{V}_{p-1}]$. 

\item For each model execution, the master decodes the groups of results $\mathcal{V} = 
[\mathcal{V}_0, \mathcal{V}_1, \dots, \mathcal{V}_{p-1}]$. For each group $\mathcal{V}_i$,
the master computes $M_i = \max(\mathcal{V}_i)$, $m_i = \min(\mathcal{V}_i),$ and the group 
diameter $D_i = M_i - m_i$. The overall maximum discrepancy across all groups is then $D = 
\max_{0 \le i \le d} D_i$. Finally, the master fixes $\tau = aD$, where $a \in \mathbb{R}$,
$a \geq 1$, is a tunable parameter for avoiding worst-case scenarios where the results 
are actually  correct.
\end{enumerate}

On top of the ADC mechanism, our scheme additionally integrates a group testing technique 
to detect Byzantine nodes, and uses a non-adaptive pooling matrix 
$\mathbf{G} \in \{ 0, 1 \}^{M \times N}$. Then, for each pool $p$ with worker set 
$S_p = \{ i: G_{p,i} = 1 \}$, the pooled reconstruction \(\widehat{\mathbf{f}}_{S_p}\) is
computed and the pool $p$ passes the test if $\widehat{\mathbf{f}}_{S_p} - \mathbf{m} \| 
\leq \tau$. Therefore, $\mathcal{P}_{\mathrm{fail}} = \{ p: \| \widehat{\mathbf{f}}_{S_p} -
\mathbf{m} \| > \tau \}$ are the failed pools, and encoding the binary pattern 
$\mathbf{y}:= \mathbb{I}(p \not\in \mathcal{P}_{\text{fail}})$ of pool outcomes, 
we just decode a suspect set $\widehat{\mathcal{A}}$ with any suitable group-testing 
decoder such as COMP, DD, or score-thresholding. Because of the approximation noise, it is
necessary to use noisy-GT decoding that tolerates false positives and false negatives.
The flagged workers $\widehat{\mathcal{A}}$ identified during the group-testing tomography
are subsequently pruned and their evaluation results of the target function are dismissed.

Finally, the desired values are reconstructed (approximately) only with the cleaned set 
$\mathcal{F}_{\mathrm{clean}}$, $\widehat{\mathbf{f}}_{\mathrm{final}} \;=\; r_{\mathrm{Berrut}, 
\mathcal{F}_{\mathrm{clean}}} (\alpha)$. Because decoding error in GPBACC decreases with 
the fraction of honest points, this prune-and-refine approach improves final precision.

The last step in the mechanism for resilience against Byzantine attackers is the 
calibration of the noisy group testing detection algorithm. To that end, let $s$ 
be the maximum expected number of adversarial workers. Choose $M = \Theta(s\log N)$
and a pooling density $\theta$ so that pools have expected size $\theta N \in 
[0.05N,  0.2N]$. Under noisy observations with false-positive rate $\alpha$ and 
false-negative rate $\beta$, the GT decoder parameters (e.g., threshold $\gamma$ 
in score decoding) must be tuned to control the trade-off between the false removal
of honest workers (reduces decoding fidelity); and the missed adversaries (leads 
to residual poisoning influence). This calibration is part of our empirical methodology 
and will be explained in Section~\ref{sec:results}.

\begin{algorithm}[t]
\caption{GT + ADC prune-and-refine}
\label{alg:gt-adc}
\begin{algorithmic}[1]
\Require Bootstrap for calculating threshold \(\tau\)
\Require Received intermediate results \(I=\{I_i\}_{i\in\mathcal{F}}\), pooling matrix \(\mathbf{G}\)
\For{each pool \(p=1,\dots,M\)}
  \State compute \(\widehat{\mathbf{f}}_{S_p}\) and set \(y_p \leftarrow \mathbb{1}[\|\widehat{\mathbf{f}}_{S_p}-\mathbf{m}\|>\tau]\)
\EndFor
\State Decode suspect set: $\widehat{\mathcal{A}} \leftarrow \mathsf{NoisyGTDecoder}(\mathbf{y})$
\State $\mathcal{F}_{\mathrm{clean}} \leftarrow \mathcal{F} \setminus \widehat{\mathcal{A}}$
\State Output final reconstruction \( \widehat{\mathbf{f}}_{\mathrm{final}} = r_{\mathrm{Berrut},\mathcal{F}_{\mathrm{clean}}}(\alpha)\)
\end{algorithmic}
\end{algorithm}

\section{Experiments and Results}
\label{sec:results}
This Section evaluates the proposed Generalized Privacy-aware Berrut Approximated 
Coded Computing scheme under both a centralized federated learning setting and a 
decentralized learning architecture, compared to the following state-of-the-art models.

For FL, GPBACC is compared to the plain unprotected version of the learning algorithm, 
and to two other techniques that rely on alternative types of PETs: an efficient Multi-Key 
Homomorphic Encryption scheme~\cite{practical-mkhe}, and the Sparse Vector 
Technique~\cite{svt} with Differential Privacy (DP-SVT). Each of these PETs was further 
tested with robust aggregation strategies, provided these are supported by the baselines: average, median, Krum and multi-Krum. The experiments have 
been conducted under diverse threat conditions: without attacks, with a label flipping 
attack,  with a Gaussian noise attack, and under a robust membership inference attack.

The case DL differs in the type of information to protect and the types of function
computed, which rendered the MKHE and SVT not applicable. For this reason, we compare the
GPBACC framework with two alternative methods: unprotected DL, and Analog Lagrange Coded 
Computing, a privacy-preserving coded computing baseline that uses analog polynomial 
encoding. Since an aggregation strategy resistant against poisoning attacks does not exist 
in DL, GPBACC is complemented with ADC + GT with refine-and-prune strategy as its 
adversary resistance countermeasure. Moreover, with ALCC,  a typical DC method was used
in combination with a very small allowed deviance, given that the reconstruction results 
of this algorithm are almost exact. All the experiments have been repeated for different 
types of attacks: no attacks, label flipping, and noise addition. The robust membership
inference attack is not reproducible under DL, so the achievable privacy has been 
measured through a membership inference estimation of the mutual information over the 
encoded datasets.

The learning models used in the experiments are convolutional neural networks over the 
same datasets (MNIST ans CIFAR10) and with the same CNN architecture, ensuring a fair 
comparison with the baseline algorithms.

\subsection{Results for Federated Learning}
We evaluate five FL configurations: unprotected FL, practical MKHE, SVT with DP, and GPBACC
(ours) with CNN+MNIST data, plus FL with a CNN+CIFAR10 under a robust membership inference
attack. Unless noted, we report numerical values for the accuracy only, since 
precision/recall/F1 follow a similar trend to accuracy. Privacy leakage is measured with 
RMIA AUC (lower is better; random $\approx 0.50$). Resilience against poisoning attacks 
is evaluated by running both label flipping and noise addition under standard aggregation
algorithms in the literature.

\begin{table}[t]
\centering
\footnotesize
\caption{Clean training on CNN--MNIST: accuracy evolution across rounds. R$X$ means average accuracy at round $X$.}
\label{tab:fl-clean-evolution}
\begin{tabular}{llccccc}
\toprule
 & \textbf{Aggr.} & \textbf{R2} & \textbf{R4} & \textbf{R6} & \textbf{R8} & \textbf{R10} \\
\midrule
\multirow{4}{*}{\textsc{fed-}} & \textsc{avg}    & 0.9559 & 0.9755 & 0.9813 & 0.9839 & 0.9855 \\
 & \textsc{median} & 0.9575 & 0.9759 & 0.9812 & 0.9838 & 0.9854 \\
 & \textsc{krum}   & 0.9397 & 0.9659 & 0.9727 & 0.9755 & 0.9767 \\
 & \textsc{mkrum}  & 0.9530 & 0.9735 & 0.9799 & 0.9832 & 0.9845 \\
\midrule
\textsc{mkhe}   & \textsc{avg}    & 0.9558 & 0.9757 & 0.9807 & 0.9838 & 0.9850 \\ \midrule
\multirow{4}{*}{\textsc{dp-svt}} & \textsc{avg}  & 0.1876 & 0.5691 & 0.6098 & 0.5827 & 0.5005 \\
 & \textsc{median} & 0.1287 & 0.1037 & 0.1046 & 0.1021 & 0.0993 \\
 & \textsc{krum} & 0.2658 & 0.3358 & 0.3054 & 0.2439 & 0.2265 \\
 & \textsc{mkrum}& 0.3013 & 0.5052 & 0.4515 & 0.4360 & 0.3571 \\ \midrule\midrule\multirow{4}{*}{\textsc{\bfseries gpbacc}} & \textsc{avg}    & 0.9563 & 0.9755 & 0.9810 & 0.9837 & 0.9853 \\
 & \textsc{median} & 0.9565 & 0.9753 & 0.9808 & 0.9835 & 0.9851 \\
 & \textsc{krum}   & 0.8831 & 0.9347 & 0.9172 & 0.9443 & 0.9497 \\
 & \textsc{mkrum}  & 0.9498 & 0.9702 & 0.9763 & 0.9806 & 0.9830 \\
\bottomrule
\end{tabular}
\end{table}

Table~\ref{tab:fl-clean-evolution} reports accuracy evolution across training rounds under 
clean (non-adversarial) conditions, raising several observations. First, the Unarmored 
baseline reaches strong performance for all aggregation rules. Simple averaging yields 
the highest accuracy at R10, while robust aggregators such as Median and 
Multi-Krum closely match this value. Krum aggregation converges more slowly 
and attains a slightly lower final accuracy due to its conservative robust selection 
mechanism. Compared to these, GPBACC preserves clean-model utility almost perfectly. 
For Average, Median and Multi-Krum, the accuracy curve remains nearly indistinguishable 
from the unarmored case, with deviations below~$4\times 10^{-4}$ at round $10$. This 
confirms that GPBACC's encoding, decoding, and interpolation introduce negligible distortion
during training. The only exception is Krum's aggregation, where GPBACC exhibits a
slower improvement over time and attains lower final accuracy. Nevertheless, this is 
expected because Krum's algorithm relies on selecting just one valid model update based 
on pairwise distances, which will be more sensitive to the small reconstruction errors 
inherent to approximate decoding. The results also show that the encryption-based MKHE,
with zero privacy leakage but a high computational cost, achieves essentially identical
performance to both GPBACC and the unarmored baseline. Finally, the use of DP as 
privacy mechanism, in SVT, severely harms clean accuracy across all aggregators. 

\begin{table}[t]
\centering
\small
\caption{RMIA AUC at round 30 (lower is better.}
\label{tab:fl-rmia-auc}
\begin{tabular}{lcccc}
\toprule
\textbf{Model} & \textbf{Uncoded} & \textbf{GPBACC} & \textbf{MKHE} & \textbf{SVT} \\
\midrule
\textsc{cnn-mnist}   & 0.5230 & 0.5077 & 0.5028 & 0.5024 \\
\textsc{cnn-cifar10} & 0.6202 & 0.5018 & 0.5022 & 0.5012 \\
\bottomrule
\end{tabular}
\end{table}

Table~\ref{tab:fl-rmia-auc} lists the AUC scores of the RMIA attacks, showing that 
protected scenarios mitigate the effects of the attack, which are visible under the 
unarmored setting. Here, CNN-CIFAR10 was also tested, given its tendency 
to overfitting, the key factor that is susceptible to MIA attacks in general.
For adapting the RMIA attack to a FL setting, the Byzantine aggregator chooses one 
data owner as victim and always sends back their received local updates to it without 
doing any aggregation, whereas it aggregates properly for the non-attacked nodes. 
This forces the training of an overfitted model having a better resilience to the 
attack for that single victim. Note that induction of overfitting is only realizable 
for the unarmored and SVT settings, since GPBACC and  MKHE secure the aggregation 
process, avoiding the ability of driving a victim to overfit. This implies that the 
attack is evaluated over the global reconstructed model in GPBACC, and from victim's 
local encrypted model in MKHE.

For the CNN–MNIST model, the unprotected system exhibits a mild but noticeable 
vulnerability, confirming that even simple models trained on low-dimensional
data can leak membership information. All three defense mechanisms substantially 
reduce the attack's effectiveness, pushing the AUC closer to random behavior 
($\text{AUC} \approx 0.5)$, with GPBACC slightly worse than MKHE and SVT, but all
essentially eliminating detectable membership signals during training.
The same pattern repeats and is even more pronounced for the CNN–CIFAR10 model, where
the unarmored configuration yields a significantly high AUC, whereas the three PET
mechanisms mitigate the attack. Importantly, the defenses not only reduce the absolute 
leakage but also normalize the privacy risk across architectures: the unprotected CIFAR10 
model leaks far more than the MNIST one, yet all defended configurations converge to nearly
identical AUC values.

\begin{table}[t]
\centering
\footnotesize
\caption{Label Flipping on CNN--MNIST: accuracy evolution (R2, R4, R6, R8, R10).}
\label{tab:fl-lf-evolution}
\begin{tabular}{llccccc}
\toprule
& \textbf{Aggr.} & \textbf{R2} & \textbf{R4} & \textbf{R6} & \textbf{R8} & \textbf{R10} \\
\midrule
\multirow{4}{*}{\textsc{fed-}} & \textsc{avg}    & 0.9531 & 0.9727 & 0.9788 & 0.9814 & 0.9829 \\
 & \textsc{median} & 0.9557 & 0.9751 & 0.9813 & 0.9837 & 0.9852 \\
 & \textsc{krum}   & 0.9416 & 0.9656 & 0.9728 & 0.9759 & 0.9764 \\
 & \textsc{mkrum}  & 0.9532 & 0.9738 & 0.9804 & 0.9829 & 0.9845 \\
\midrule
\textsc{mkhe}   & \textsc{avg}    & 0.9526 & 0.9696 & 0.9772 & 0.9813 & 0.9830 \\ \midrule
\multirow{4}{*}{\textsc{dp-svt}} & \textsc{avg}  & 0.1493 & 0.5311 & 0.5901 & 0.5053 & 0.4618 \\
 & \textsc{median} & 0.1195 & 0.0990 & 0.1021 & 0.1129 & 0.1016 \\
 & \textsc{krum} & 0.2198 & 0.2797 & 0.2753 & 0.2486 & 0.2333 \\
 & \textsc{mkrum}& 0.2203 & 0.4620 & 0.4822 & 0.4111 & 0.3556 \\ \midrule\midrule\multirow{3}{*}{\textsc{\bfseries gpbacc}} & \textsc{avg}    & 0.9521 & 0.9719 & 0.9785 & 0.9814 & 0.9831 \\
 & \textsc{median} & 0.9532 & 0.9722 & 0.9785 & 0.9813 & 0.9831 \\
 & \textsc{krum}   & 0.8715 & 0.9155 & 0.9250 & 0.9135 & 0.9720 \\
 & \textsc{mkrum}  & 0.9372 & 0.9641 & 0.9701 & 0.9721 & 0.9720 \\
\bottomrule
\end{tabular}
\end{table}

Table~\ref{tab:fl-lf-evolution} examines accuracy  under a label-flipping attack. 
In the unprotected setting (the reference baseline), the label 
flipping attack does not have a strong effect on the overall model convergence. 
It mainly affects the first round of the protocol, and it then slightly reduces 
the maximum accuracy achieved for the standard federated averaging mechanism. 
Krum's aggregation consistently underperforms the others.
For GPBACC, the results closely mirror this baseline under average and median 
aggregation, reinforcing that GPBACC’s privacy mechanism does not hinder model
convergence while still providing privacy protection.  The most interesting behavior 
arises from GPBACC + Krum, which displays a non-monotonic and noticeably lower accuracy 
profile. Accuracy drops sharply compared to all other configurations, though raises again
at R10. This suggests an interaction effect: the additional variance and structure 
introduced by GPBACC encoding can skew Krum’s distance-based scoring, causing it to 
select suboptimal updates, especially in early rounds where gradient signal-to-noise 
ratios are already low. Nevertheless, the recovery by R10 indicates that the model still 
converges once gradients stabilize. MKHE matches the baseline almost exactly, indicating 
that homomorphic encryption does not distort gradient distributions enough to weaken 
robustness.  DP-SVT fails to converge, providing no utility for learning.

\begin{table}[t]
\centering
\footnotesize
\caption{Noise Addition on CNN--MNIST: accuracy evolution (R2, R4, R6, R8, R10).}
\label{tab:fl-noise-evolution}
\begin{tabular}{llccccc}
\toprule
 & \textbf{Aggr.} & \textbf{R2} & \textbf{R4} & \textbf{R6} & \textbf{R8} & \textbf{R10} \\
\midrule
\multirow{4}{*}{\textsc{fed-}} & \textsc{avg}    & 0.3873 & 0.4775 & 0.5238 & 0.5860 & 0.6338 \\
 & \textsc{median} & 0.9546 & 0.9744 & 0.9806 & 0.9831 & 0.9847 \\
 & \textsc{krum}   & 0.9418 & 0.9662 & 0.9712 & 0.9720 & 0.9743 \\
 & \textsc{mkrum}  & 0.9537 & 0.9736 & 0.9792 & 0.9819 & 0.9837 \\
\midrule
\textsc{mkhe}   & \textsc{avg}    & 0.2659 & 0.4000 & 0.4944 & 0.5637 & 0.6197 \\ \midrule
\multirow{4}{*}{\textsc{dp-svt}} & \textsc{avg}  & 0.1876 & 0.5691 & 0.6098 & 0.5827 & 0.5005 \\
 & \textsc{median} & 0.1287 & 0.1037 & 0.1046 & 0.1021 & 0.0993 \\
 & \textsc{krum} & 0.2658 & 0.3358 & 0.3054 & 0.2439 & 0.2265 \\
 & \textsc{mkrum} & 0.3013 & 0.5052 & 0.4515 & 0.4360 & 0.3571 \\ \midrule
 \midrule
 \multirow{4}{*}{\textsc{\bfseries gpbacc}} & \textsc{avg}    & 0.3749 & 0.4614 & 0.5374 & 0.5640 & 0.5901 \\
 & \textsc{median} & 0.9336 & 0.9598 & 0.9669 & 0.9737 & 0.9759 \\
 & \textsc{krum}   & 0.8848 & 0.9173 & 0.9548 & 0.9608 & 0.9587 \\
 & \textsc{mkrum}  & 0.9533 & 0.9728 & 0.9787 & 0.9817 & 0.9834 \\
\bottomrule
\end{tabular}
\end{table}

Table~\ref{tab:fl-noise-evolution} evaluates the impact of Gaussian noise injection on
CNN–MNIST training. Unlike label flipping, where malicious gradients are structured,
adversarial—noise addition primarily stresses the stability of aggregators and their
ability to extract signal from highly perturbed updates. In the non-private unprotected, 
the contrast between aggregation rules is immediately evident. FedAvg fails, as injected 
noise directly corrupts the global update, but Median, Krum, and multi-Krum all attain
strong resilience, confirming as in other works that robust aggregation is essential
when gradient reliability is low. Under private aggregation, only our GPBACC is capable 
of keeping nearly the same performance as with non-private robust aggregation, while 
MKHE and differential privacy are completely ineffective.

The noise addition experiment highlights three key insights. First, robust
aggregators are indispensable when clients submit highly noisy
updates, with median providing the greatest stability. Second, GPBACC and MKHE do not
meaningfully degrade robustness; the determining factor remains the aggregation strategy
supported rather than the PET mechanism. Third, DPSVT imposes the steepest utility cost, and
its instability under noise suggests that combining heavy DP noise with adversarial or
unreliable updates can lead to degradation.

\subsection{Results for Decentralized Learning}
\label{sec:decentralized}

In a decentralized learning network, three configurations are compared for a CNN+MNIST
model: no privacy protection (baseline), ALCC, and GPBACC (ours).
Again, we report just accuracy since precision/recall/F1 follow the same pattern of
variation as the accuracy values. For the quantification of privacy, RMIA is no longer 
a valid strategy, since in DL local model updates are not exchanged, only model execution
results. In this case, in practice, we measure the mutual information of the encoded 
datasets, and compare them with the theoretical bounds, validating their privacy 
guarantees provided. Similar observations hold with poisoning attacks, in that label 
flipping cannot be applied anymore because workers do not have access to labels, so 
noise addition over the model execution results is the only possible threat to 
demonstrate practical privacy.

GPBACC uses ADC+GT technique for resisting adversary behaviour while we implemented vanilla
Decode and Compare for ALCC (with a very small error bound), since ALCC provides almost
perfect results. Additionally, since ALCC cannot directly compute a CNN, we used polynomial
approximations for the activation functions of the model architecture. Please note that this
has an impact on precision that highly variates depending on the model trained, even making
some models unfeasible to use. Additionally, this also adds an additional overhead in the
researching task, since the most popular ML frameworks do not support these approximations
natively, and they must be explictly created. Finally, there is also an exchange between grade
of the approximated model and the amount of adversary nodes you can resist using ALCC, since
more grade requires more correct results to reconstruct.

\begin{table}[t]
\centering
\small
\caption{Clean decentralized learning over CNN--MNIST: accuracy evolution across training rounds.}
\label{tab:dl-clean-dec}
\begin{tabular}{lccccc}
\toprule
 & \textbf{R1} & \textbf{R2} & \textbf{R3} & \textbf{R4} & \textbf{R5} \\
\midrule
\textsc{dfl} & 0.9643 & 0.9723 & 0.9675 & 0.9733 & 0.9808 \\
\textsc{alcc}      & 0.9645 & 0.9684 & 0.9743 & 0.9707 & 0.9772 \\
\textsc{\bfseries gpbacc}    & $\mathbf{0.9648}$ & $\mathbf{0.9765}$ & $0.9658$ & $\mathbf{0.9801}$ & $0.9716$ \\
\bottomrule
\end{tabular}
\end{table}
Table~\ref{tab:dl-clean-dec} reports the accuracy evolution across training rounds for clean
\emph{decentralized learning} on CNN--MNIST. In the absence of adversarial behavior, all methods
achieve comparable performance, indicating that neither coding nor verification mechanisms inherently
degrade learning under benign conditions.

The Unarmored baseline shows stable convergence, reaching an accuracy of $0.9808$ at R5.
ALCC closely tracks this trajectory, with marginal round-to-round fluctuations. This confirms
that exact coded computing does not introduce noticeable optimization bias in decentralized settings.

GPBACC exhibits similar average performance but with increased variance across rounds, while it
attains a similar highest accuracy at R4 ($0.9801$). This behavior is consistent with GPBACC’s
approximate reconstruction error, which introduces small numerical perturbations that barely affects
convergence in fully decentralized optimization.

\begin{table}[t]
\centering
\small
\caption{Decentralized learning under noise addition attack: accuracy evolution across training rounds.}
\label{tab:noise-dec}
\begin{tabular}{lccccc}
\toprule
 & \textbf{R1} & \textbf{R2} & \textbf{R3} & \textbf{R4} & \textbf{R5} \\
\midrule
\textsc{dfl} & 0.6151 & 0.6237 & 0.6320 & 0.6338 & 0.6326 \\
\textsc{alcc}      & 0.9610 & 0.9653 & 0.9767 & 0.9764 & 0.9741 \\
\textsc{\bfseries gpbacc}    & $\mathbf{0.9692}$ & $\mathbf{0.9683}$ & 0.9719 & 0.9636 & $\mathbf{0.9779}$ \\
\bottomrule
\end{tabular}
\end{table}

Table~\ref{tab:noise-dec} reports the accuracy evolution under a noise addition attack in 
decentralized learning. In this setting, a subset of workers injects additive noise into 
their local updates, directly corrupting intermediate results exchanged with the master.

The Unarmored baseline is highly vulnerable to this attack. Accuracy collapses to 
approximately $0.63$ from the first round and remains stagnant across subsequent rounds,
indicating that even moderate noise severely disrupts convergence when no protection 
mechanism is in place. In contrast, both GPBACC and ALCC maintain high accuracy throughout 
training. ALCC achieves stable performance, converging around $0.97 \sim 0.98$ by R5, 
confirming the effectiveness of exact coded redundancy in filtering noisy contributions. 
GPBACC performs comparably, with accuracy consistently above $0.96$. Minor round-to-round
fluctuations are expected due to GPBACC’s approximate reconstruction, but these do not 
hinder overall convergence.

Overall, these results demonstrate that coded computing mechanisms are essential for 
decentralized learning robustness. GPBACC achieves noise resilience comparable to exact 
ALCC while being more flexible in distributed learning strategies and model architectures,
highlighting its suitability for adversarial decentralized settings where both robustness and 
privacy are required.

\subsection{Privacy and Data Obfuscation}
\label{sec:privacy_dec}

In decentralized learning configuration, the master is the owner of the data, which is 
exchanged encoded to the workers. This makes RMIA unusable for validating the practical 
privacy of this encoded dataset. Therefore, we will leverage a numerical estimator of the 
mutual information between the plain dataset ($X$) and the encoded one ($Y$), and 
compare this result with the theoretical privacy leakage metric~\eqref{eq:il-normalized}.

For this practical measure, we adopt the KSG estimator~\cite{ksg-mi-estimator}. This 
estimator is based on the observation that mutual information can de defined as a
difference of entropies, $I(X;Y) = H(X) + H(Y) - H(X,Y)$, and that each entropy term can 
be estimated from local neighborhood statistics obtained via $k$-nearest neighbors. 
Crucially, KSG enforces a shared local neighborhood scale across joint and marginal 
spaces, which significantly reduces bias compared to naive entropy-difference estimators.
For each sample $(x_i,y_i)$, distances in joint space are computed using the $\ell_\infty$
norm:
\begin{equation*}
\|(x_i,y_i)-(x_j,y_j)\|_\infty
= \max\big(\|x_i-x_j\|, \|y_i-y_j\|\big).
\end{equation*}

Let $\epsilon_i$ denote the distance from $(x_i,y_i)$ to its $k$-th nearest neighbor in joint
space under this metric, and
\begin{align*}
n_x(i) &= \#\{j \neq i : \|x_i - x_j\| < \epsilon_i\}, \\
n_y(i) &= \#\{j \neq i : \|y_i - y_j\| < \epsilon_i\}.
\end{align*}
The KSG estimator of mutual information is given by
\begin{equation*}
\widehat{I}_{\text{KSG}}
=
\psi(k) + \psi(N)
- \frac{1}{N}\sum_{i=1}^N
\left[
\psi(n_x(i)+1) + \psi(n_y(i)+1)
\right],
\end{equation*}
where $\psi(\cdot)$ denotes the digamma function.

After applying this numerical estimator, we obtain a value of $\widehat{I}_{\text{KSG}} = 1.012$ bits,
which is very similar to the one obtained when applying the theoretic privacy metric ($i_L\leq 1$ bit).
This implies that the assumptions we have made to bound the theoretic privacy are reasonable, and the
metric values are quite close to reality, at least for this type of datasets.


\section{Conclusions}
\label{sec:conclusiones}

In this paper, we introduce a unified framework for adversary-resistant distributed machine learning that jointly addresses privacy leakage and enhances resilience against active malicious adversaries in both federated and decentralized learning paradigms. In contrast to prior work in this field, which typically addresses privacy protection, robust aggregation, or verification mechanisms in isolation, our approach defines a solution that explicitly evaluates privacy leakage and malicious behavior under realistic attack scenarios.

A key contribution of our research work is the explicit, attack-driven evaluation of both privacy and malicious resistance. Rather than assuming protection by construction, we implemented representative privacy attacks and adversarial behaviors, and evaluated their effectiveness against the proposed framework. The results show that the combination of approximate computations and a secret sharing scheme with robust aggregation and verification mechanisms significantly reduces privacy leakage while maintaining resilience against active adversaries. Consequently, our approach provides empirical evidence that GPBACC, when combined with appropriate adversary-resistance mechanisms, can offer a practical and deployable alternative for real-world distributed learning systems. 
We are currently working on extending our proposal to more complex threat models and learning architectures, particularly those relevant to generative AI frameworks, given their pervasive relevance.

\bibliographystyle{IEEEtran}
\bibliography{references}  

\end{document}